\documentclass{article}
\usepackage{spconf,amsmath,graphicx}

\usepackage{enumitem}
\setlist{nosep, leftmargin=14pt}

\usepackage{mwe} 
\usepackage{xcolor} 

\title{Improving PI-RADS prediction and prostate cancer detection from multiparametric MRI with anatomical guidance}

\name{
\shortstack{
Seungbin Park$^{\star}$, Thibault Heintz$^{\star}$, Rachit Saluja$^{\dagger\star}$, Francesca Khani$^{\star}$, \\
Timothy McClure$^{\star}$, Daniel Margolis$^{\star}$, Mert R. Sabuncu$^{\star\dagger}$, Heejong Kim$^{\star\ddagger}$}
\thanks{$^{\ddagger}$Corresponding author. hek4004@med.cornell.edu}}

\address{
$^{\star}$ Weill Cornell Medicine, New York, NY, USA \\
$^{\dagger}$ Cornell Tech, New York, NY, USA
}
\usepackage{booktabs}
\usepackage{multirow} 
\usepackage{xcolor}
\usepackage{comment}
\usepackage{graphicx}
\usepackage[normalem]{ulem}
\usepackage[table]{xcolor}
\usepackage{hhline}
\usepackage{hyperref}

\begin{document}
%
\maketitle

\begin{abstract}
Accurate prediction of Prostate Imaging Reporting and Data System (PI-RADS) scores from multiparametric MRI (mpMRI) remains challenging, more so than clinically significant prostate cancer (csPCa) detection. 
Given the importance of lesion location in PI-RADS scoring, we evaluated whether incorporating prostate zonal masks improves PI-RADS prediction and csPCa detection across multiple architectures (3D ResNet-18, ResNet-50, and ViT) using a large-scale internal dataset.
Anatomical guidance generally improved performance across architectures, with the best results achieved by ResNet-18 with 2-zone guidance for PI-RADS prediction (QWK = 0.511) and ResNet-50 with 5-zone guidance for csPCa detection (AUC = 0.809). Fine-tuned results on external datasets further demonstrated the benefits of anatomical guidance.
Furthermore, models with anatomical guidance improved the utility of model-predicted PI-RADS scores for PI-RADS-based csPCa detection (positive defined as PI-RADS$\geq$3).
Combining model-predicted and expert-assigned PI-RADS scores significantly increased the AUC from 0.711 to 0.756 ($p<0.05$), suggesting that AI-derived PI-RADS scores may provide complementary information to expert assessment for csPCa detection.
Overall, our findings highlight the value of incorporating prostate zonal anatomy into deep learning-based prostate MRI assessment.

\end{abstract}
\section{Introduction}
\label{sec:intro}

\begin{figure}[t!]
    \centering
    \includegraphics[width=0.95\columnwidth]{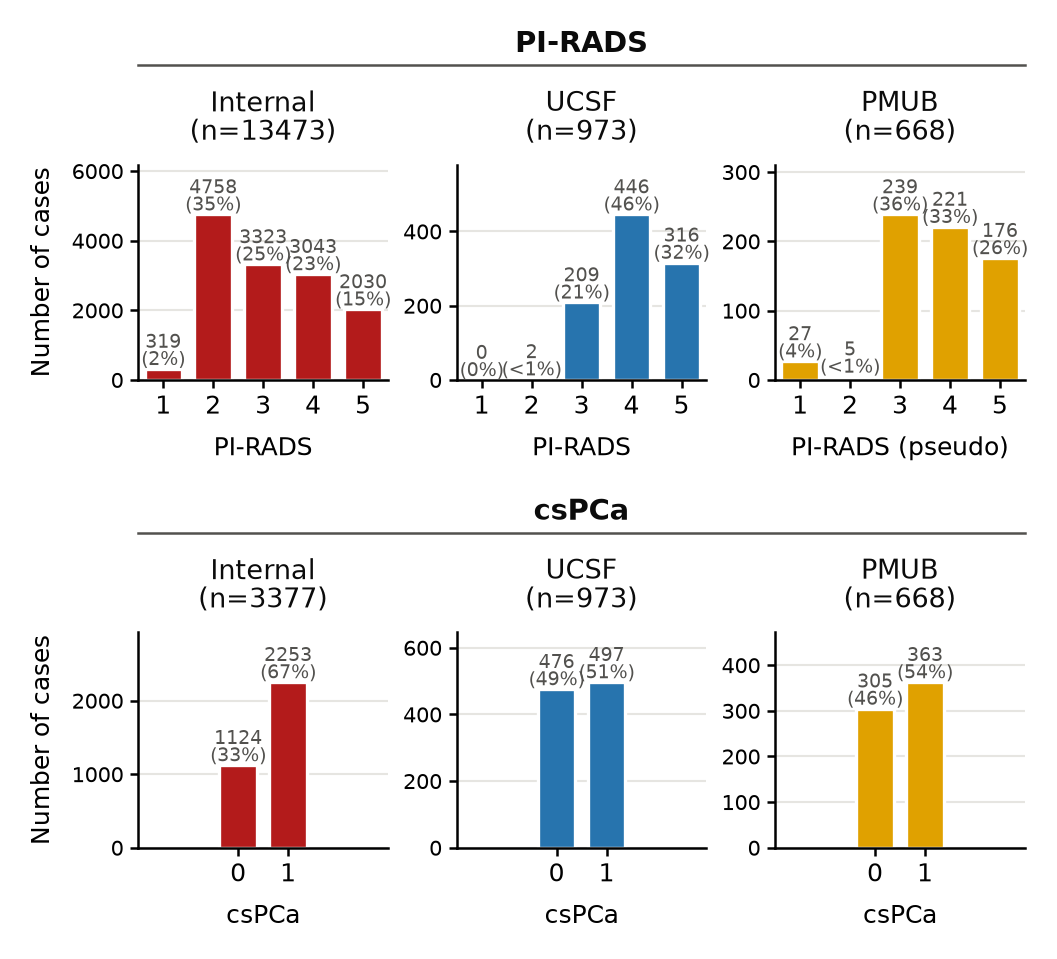}
    \caption{Distribution of PI-RADS and csPCa labels across the internal, UCSF, and PMUB datasets.}
    \label{fig:distribution}
\end{figure}

Multiparametric magnetic resonance imaging (mpMRI) is widely used to identify suspicious prostate lesions and guide biopsy decisions. 
The Prostate Imaging Reporting and Data System (PI-RADS) \cite{turkbey_prostate_2019} standardizes mpMRI interpretation by assigning scores from 1 to 5, with a score of 5 indicating clinically significant prostate cancer (csPCa) is highly likely.
In clinical practice, PI-RADS$\geq$3 is a common threshold for biopsy referral, and subsequent histopathological evaluation provides the definitive csPCa diagnosis \cite{padhani_pi-rads_2019}.

Accurate PI-RADS assessment is essential for biopsy decisions.
Compared with csPCa detection, automated PI-RADS prediction from mpMRI has received relatively limited attention. 
Existing deep learning (DL) approaches have demonstrated modest predictive performance \cite{sanford_deeplearningbased_2020, mehralivand_cascaded_2022, wang_profound_2026}.
Challenges stem from substantial inter-reader variability \cite{rosenkrantz_interobserver_2016}, overlapping imaging characteristics of benign and malignant lesions, and imbalanced PI-RADS score distributions \cite{pooch_critical_2026}.
Publicly available prostate MRI datasets are relatively small and have highly imbalanced PI-RADS score distributions, as they primarily include patients who underwent biopsy \cite{pooch_critical_2026} (Fig.~\ref{fig:distribution}).
To overcome the limitations of existing datasets, we trained and evaluated DL models on a large-scale in-house cohort of $13,473$ mpMRI examinations with patient-level PI-RADS scores.

\begin{table}[!t]
\centering
\caption{Summary of dataset sizes across training, validation, and test sets. Adjusted csPCa labels treat PI-RADS 1 and 2 cases as negative. Pseudo PI-RADS is radiologist-assigned suspicion scores which approximately correspond to PI-RADS v2 \cite{natarajan_prostate_2020}.}
\label{tab:dataset_distribution}
\vspace{6pt}
\resizebox{\columnwidth}{!}{%
\begin{tabular}{llrrrr}
\hline
Dataset & Task & Train & Val & Test & Total \\
\hline

\multirow{2}{*}{Internal}
& PI-RADS & 10,852 & 688 & 1,933 & 13,473 \\
& csPCa (adjusted) & 6,491 & 437 & 1,257 & 8,185 \\

\hline

\multirow{2}{*}{UCSF}
& PI-RADS & 678 & 98 & 195 & 971 \\
& csPCa & 680 & 98 & 195 & 973 \\

\hline

\multirow{2}{*}{PMUB}
& PI-RADS (pseudo) & 159 & 95 & 382 & 636 \\
& csPCa & 167 & 100 & 401 & 668 \\

\hline
\end{tabular}}
\end{table}

Prostate zonal anatomy determines how a lesion is assessed in PI-RADS. Diffusion-weighted imaging (DWI) is the dominant sequence for peripheral zone (PZ) lesions, whereas T2-weighted (T2W) imaging is dominant in the transition zone (TZ)\cite{turkbey_prostate_2019}.
Radiologists interpret imaging findings in the context of zonal anatomy, yet most DL models learn directly from mpMRI without such guidance.
Previous studies have shown that incorporating anatomical guidance can improve prostate cancer detection and lesion characterization \cite{zheng_atpca-net_2024, karagoz_anatomically_2023, hosseinzadeh_deep_2022}.
However, the impact of anatomical guidance on PI-RADS prediction remains underexplored.

In this work, we incorporate prostate zone masks as anatomical guidance into DL models and systematically evaluate its effect on PI-RADS and csPCa prediction.
We evaluate 3D Vision Transformer (ViT), ResNet-18, and ResNet-50 architectures with and without 2-zone or 5-zone anatomical guidance using internal dataset and two external datasets. 
We further assessed the clinical utility of model-predicted PI-RADS scores by comparing model-predicted and expert-assigned PI-RADS scores for PI-RADS-based csPCa detection (positive defined as PI-RADS$\geq$3).
Our results demonstrate that anatomical guidance consistently improves model performance with generalizability across datasets.
We publicly release the training code (https://github.com/seungbin20
1803/prostate-anatomy-guidance-ai) and model weights (https:
//huggingface.co/spark-25/prostate-anatomy-guidance-ai).

\section{Methods and Experiments}
\label{sec:format}

\subsection{Datasets}
\label{ssec:Data}
We used an internal dataset and two external datasets: UCSF-ProstateMR (UCSF) \cite{rajagopal_mixed_2024} and Prostate-MRI-US-Biopsy (PM
UB) \cite{natarajan_prostate_2020}.
Datasets include mpMRI sequences comprising T2W, DWI, and apparent diffusion coefficient (ADC) maps, along with patient-level PI-RADS scores and csPCa labels.
For csPCa prediction, analyses were restricted to cases with biopsy-confirmed outcomes. 
Dataset sizes and split distributions are summarized in Table \ref{tab:dataset_distribution}.
All images across datasets were resampled to $1 \times 1 \times 3$ mm$^3$ and cropped to a volume of $112 \times 112 \times 24$ voxels centered on the prostate using the prostate masks provided in the datasets.
For ProFound (see \ref{ssec:Experiments}), images were resampled to $0.5 \times 0.5 \times 1$ mm$^3$ and cropped to $224 \times 224 \times 64$ voxels, to match the pretrained encoder.

\subsection{Zonal Segmentation}
\label{ssec:Segmentation}
\begin{figure}[!t]
    \centering
    \includegraphics[width=0.8\columnwidth,
    keepaspectratio]{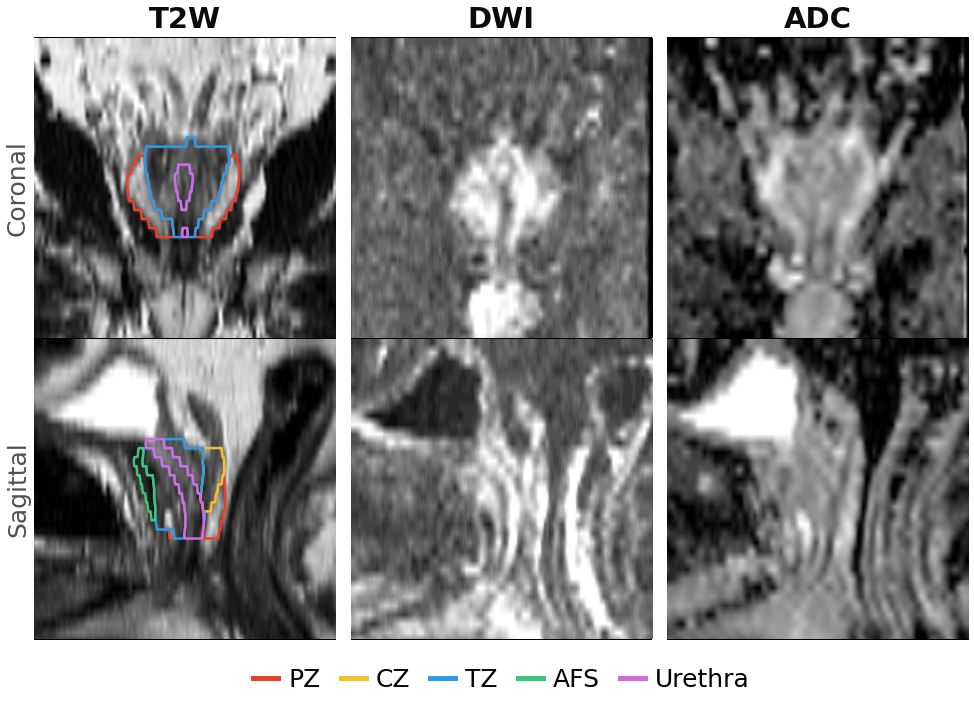}
    \caption{Representative T2W, DWI, and ADC images with corresponding five-zone prostate segmentation masks. PZ: peripheral zone; CZ: central zone; TZ: transition zone; AFS: anterior fibromuscular stroma.}
    \label{fig:data_example_slices}
\end{figure}

We constructed zonal masks of PZ, central zone (CZ), TZ, anterior fibromuscular stroma (AFS), and urethra from T2W using a publicly available pretrained 3D nnU-Net \cite{holmlund_prostatezones_2024}. 
The 5-zone guidance includes all five anatomical regions as separate masks, whereas the 2-zone configuration combined the CZ, TZ, AFS, and urethra into a single TZ mask while preserving the PZ mask. Fig. \ref{fig:data_example_slices} shows T2W, DWI, and ADC images, along with prostate five-zone segmentation masks.

\subsection{Model Development}
\label{ssec:training}
We trained 3D ViT~\cite{dosovitskiy2021image}, ResNet-18, and ResNet-50 \cite{he2016deep} models with and without prostate zonal masks.
T2W, DWI, and ADC volumes were concatenated along the channel dimension, and zonal masks were appended as additional channels when anatomical guidance was used.

PI-RADS prediction was formulated as a five-class ordinal classification task using a CORAL head and loss~\cite{coral2020} to account for the inherent ordering of PI-RADS scores. For csPCa detection, models were trained as binary classifiers using cross-entropy loss. 
All models were optimized with AdamW using learning rates of $2 \times 10^{-4}$ for models and learning rate $1 \times 10^{-3}$ for the classification layers, weight decay of $1 \times 10^{-5}$, and cosine learning rate schedule.
Batch size was 4 with 16 gradient accumulation steps, yielding an effective batch size of 64.

Models were trained for up to 200 epochs with early stopping if the validation metric failed to improve by at least 0.005 for 15 consecutive epochs. Model selection was based on quadratic weighted kappa (QWK) for PI-RADS prediction and area under the receiver operating characteristic curve (AUC) for csPCa detection.

\begin{table}[!t]
\centering
\caption{Performance for PI-RADS prediction and csPCa detection on the internal dataset.
Values are reported as point estimates with 95\% confidence intervals.
Underlined values indicate the best-performing model.
Statistical significance was assessed against the corresponding model without anatomical guidance (gray) using paired tests ($^{*}p<0.05$).}
\label{tab:internal}

\setlength{\aboverulesep}{0pt}
\setlength{\belowrulesep}{0pt}
\setlength{\cmidrulesep}{0pt}
\arrayrulecolor{black}
\setlength{\arrayrulewidth}{0.4pt}

\vspace{6pt}

\resizebox{\columnwidth}{!}{%
\begin{tabular}{lcc}
\hline
& \multicolumn{1}{c}{\shortstack{\rule{0pt}{9pt}PI-RADS\\(Ordinal multi-label; \\ 1, 2, 3, 4, 5)}}
& \multicolumn{1}{c}{\shortstack{csPCa\\(Binary-label; \\ Positive/Negative)}} \\
\cmidrule(r{4pt}){2-2}\cmidrule(l{4pt}){3-3}
& QWK & AUC \\
\hline

ProFound-Conv
& 0.181 (0.141, 0.217)
& 0.710 (0.680, 0.741) \\

\hline
\noalign{\vskip 1pt}

\cellcolor{gray!15}3D ViT
& \cellcolor{gray!15}0.182 (0.140, 0.225)
& \cellcolor{gray!15}0.636 (0.604, 0.669) \\

3D ViT+2zones
& 0.215 (0.172, 0.257)
& 0.678 (0.648, 0.708)$^{*}$ \\

3D ViT+5zones
& \textbf{0.232 (0.190, 0.272)}$^{*}$
& \textbf{0.689 (0.659, 0.719)}$^{*}$ \\

\hline
\noalign{\vskip 1pt}

\cellcolor{gray!15}3D ResNet-18
& \cellcolor{gray!15}0.471 (0.434, 0.507)
& \cellcolor{gray!15}0.789 (0.761, 0.816) \\

3D ResNet-18+2zones
& \uline{\textbf{0.511 (0.474, 0.544)}}$^{*}$
& \textbf{0.797 (0.770, 0.823)} \\

3D ResNet-18+5zones
& 0.488 (0.449, 0.524)
& 0.787 (0.759, 0.817) \\

\hline
\noalign{\vskip 1pt}

\cellcolor{gray!15}3D ResNet-50
& \cellcolor{gray!15}0.453 (0.416, 0.489)
& \cellcolor{gray!15}0.781 (0.752, 0.809) \\

3D ResNet-50+2zones
& \textbf{0.498 (0.460, 0.532)}$^{*}$
& 0.804 (0.777, 0.831)$^{*}$ \\

3D ResNet-50+5zones
& 0.482 (0.446, 0.519)$^{*}$
& \uline{\textbf{0.809 (0.782, 0.834)}}$^{*}$ \\

\hline
\end{tabular}%
}
\end{table}

\subsection{Experimental Design}
\label{ssec:Experiments}
We first evaluated the effect of 2-zone and 5-zone guidance across ViT, ResNet-18, and ResNet-50 architectures for both PI-RADS prediction and csPCa detection, on the internal dataset.
To benchmark our approach against a state-of-the-art method, we compared our approach with ProFound \cite{wang_profound_2026}, a foundation model for prostate MRI. Publicly available pretrained encoder weights were used, and only the classification layers were fine-tuned.

To assess generalizability, models pretrained on the internal data were subsequently fine-tuned and evaluated on two external datasets (UCSF and PMUB).
PI-RADS 1 and 2 cases were excluded because of their small sample sizes.
During fine-tuning, pretrained models and randomly initialized classification layers to account for the restricted PI-RADS score range in the external datasets were optimized. 
Fine-tuning was performed for up to 50 epochs using learning rates of $2 \times 10^{-5}$ and $1 \times 10^{-3}$, respectively, with a weight decay of $1 \times 10^{-2}$. The learning rates were linearly warmed up over the first two epochs, followed by cosine decay. 
Early stopping was applied with a patience of 15 epochs and a minimum validation improvement of 0.01, after an initial 10-epoch training period.

We also evaluated the clinical utility of model-predicted PI-RADS scores for PI-RADS-based csPCa detection, defining PI-RADS $\geq$3 as a positive.
Expert-assigned PI-RADS scores, model-predicted PI-RADS scores, and their ensemble were compared for PI-RADS-based csPCa detection.
For ensembling, expert-assigned PI-RADS scores were linearly rescaled to [0, 1] using (PI-RADS - 1)/4 and averaged with the model-predicted probability of PI-RADS $\geq$ 3 to obtain a combined score.

\section{Results}
\label{sec:Results}
\subsection{Internal Evaluation}
Table \ref{tab:internal} shows that incorporating anatomical guidance generally improved both PI-RADS prediction and csPCa detection across architectures. 
For ViT, 5-zone guidance significantly improved both PI-RADS prediction (QWK: 0.182 to 0.232) and csPCa detection (AUC: 0.636 to 0.689; both $p<0.05$).
Similarly, 2-zone guidance significantly improved PI-RADS prediction for both ResNet-18 (QWK: 0.471 to 0.511) and ResNet-50 (QWK: 0.453 to 0.498; $p<0.05$).
For csPCa detection, ResNet-50 benefited from both 2-zone guidance (AUC: 0.781 to 0.804; $p<0.05$) and 5-zone guidance (AUC: 0.809; $p<0.05$). 
Overall, ResNet-18 with 2-zone guidance achieved the highest PI-RADS prediction performance (QWK = 0.511), whereas ResNet-50 with 5-zone guidance achieved the highest csPCa detection performance (AUC = 0.809).

\begin{table}[!t]
\centering
\caption{Performance on external datasets of models trained from scratch or fine-tuned from models pretrained on the internal dataset.
Values are point estimates with 95\% confidence intervals.
Statistical significance was tested against the corresponding model without anatomical guidance (gray) using paired tests ($^{*}p<0.05$).
}
\label{tab:external}

\setlength{\aboverulesep}{0pt}
\setlength{\belowrulesep}{0pt}
\setlength{\cmidrulesep}{0pt}
\setlength{\arrayrulewidth}{0.4pt}
\arrayrulecolor{black}

\vspace{6pt}

\resizebox{\columnwidth}{!}{%
\begin{tabular}{llcc}
\hline
& & \multicolumn{1}{c}{\shortstack{\rule{0pt}{9pt}PI-RADS\\(Ordinal multi-label; \\ 3, 4, 5)}}
& \multicolumn{1}{c}{\shortstack{csPCa\\(Binary-label; \\ Positive/Negative)}} \\
\cmidrule(r{4pt}){3-3}\cmidrule(l{4pt}){4-4}
& & QWK & AUC \\
\hline
\noalign{\vskip 1pt}

\multirow{8}{*}{UCSF}
& 3D ResNet-18, scratch
& 0.071 (-0.045, 0.184)
& 0.503 (0.416, 0.580) \\

& \cellcolor{gray!15}3D ResNet-18
& \cellcolor{gray!15}0.152 (0.020, 0.280)
& \cellcolor{gray!15}0.633 (0.552, 0.706) \\

& 3D ResNet-18+2zones
& \textbf{0.167 (0.032, 0.291)}
& \uline{\textbf{0.668 (0.586, 0.746)}} \\

& 3D ResNet-18+5zones
& 0.125 (-0.007, 0.253)
& 0.624 (0.541, 0.697) \\

\cline{2-4}
\noalign{\vskip 1pt}

& 3D ResNet-50, scratch
& 0.053 (-0.064, 0.161)
& 0.590 (0.509, 0.671) \\

& \cellcolor{gray!15}3D ResNet-50
& \cellcolor{gray!15}-0.005 (-0.133, 0.140)
& \cellcolor{gray!15}0.604 (0.522, 0.679) \\

& 3D ResNet-50+2zones
& 0.134 (0.017, 0.252)
& \textbf{0.611 (0.533, 0.689)} \\

& 3D ResNet-50+5zones
& \uline{\textbf{0.231 (0.116, 0.336)}}$^{*}$
& 0.605 (0.520, 0.681) \\

\hline
\noalign{\vskip 1pt}

\multirow{8}{*}{PMUB}
& 3D ResNet-18, scratch
& 0.065 (-0.020, 0.160)
& 0.519 (0.462, 0.576) \\

& \cellcolor{gray!15}3D ResNet-18
& \cellcolor{gray!15}0.296 (0.204, 0.391)
& \cellcolor{gray!15}0.659 (0.609, 0.711) \\

& 3D ResNet-18+2zones
& \uline{\textbf{0.368 (0.279, 0.454)}}
& 0.668 (0.616, 0.721) \\

& 3D ResNet-18+5zones
& 0.251 (0.154, 0.340)
& \uline{\textbf{0.683 (0.631, 0.733)}} \\

\cline{2-4}
\noalign{\vskip 1pt}

& 3D ResNet-50, scratch
& 0.022 (-0.072, 0.127)
& 0.585 (0.526, 0.642) \\

& \cellcolor{gray!15}3D ResNet-50
& \cellcolor{gray!15}0.282 (0.191, 0.374)
& \cellcolor{gray!15}0.594 (0.541, 0.648) \\

& 3D ResNet-50+2zones
& 0.356 (0.259, 0.448)
& 0.586 (0.533, 0.641) \\

& 3D ResNet-50+5zones
& \textbf{0.365 (0.280, 0.451)}
& \textbf{0.640 (0.586, 0.692)} \\

\hline
\end{tabular}%
}
\end{table}

Error analysis suggested that model performance was driven primarily by the reference labels.
For ResNet-18
+2zones, PI-RADS 1 and 4 cases were rarely predicted correctly, and csPCa-positive cases were substantially more challenging than negative ones (39\% vs. 95\%).
Neither patient age, reporting radiologist, nor examination type (initial versus follow-up) was significantly associated with performance. 
Notably, correct PI-RADS prediction strongly correlated with accurate csPCa detection (88\% vs. 65\%; odds ratio = 4.04, $p < 0.001$).

\begin{table}[!t]
\centering
\caption{Comparison of PI-RADS-based csPCa detection (PI-RADS$\geq$3 considered positive) performance using model-predicted, expert-assigned, and combined PI-RADS scores, along with direct csPCa detection by the model shown in the last row.
Values are point estimates with 95\% confidence intervals.
Statistical significance was assessed against expert scores (gray) using two-sided DeLong tests ($^{*}p<0.05$).}

\vspace{6pt}

\label{tab:pirad-cspca}

\resizebox{0.7\columnwidth}{!}{%
\begin{tabular}{lc}
\hline
& AUC \\
\hline

ProFound-Conv
& 0.597 (0.547, 0.641)$^{*}$ \\

ViT
& 0.588 (0.539, 0.634)$^{*}$ \\

ViT+2zones
& 0.639 (0.593, 0.679)$^{*}$ \\

ViT+5zones
& 0.653 (0.607, 0.694)$^{*}$ \\

ResNet-18
& 0.678 (0.637, 0.717) \\

ResNet-18+2zones
& 0.710 (0.668, 0.750) \\

ResNet-18+5zones
& 0.727 (0.688, 0.764) \\

ResNet-50
& 0.704 (0.661, 0.742) \\

ResNet-50+2zones
& \textbf{0.729 (0.689, 0.771)} \\

ResNet-50+5zones
& 0.708 (0.668, 0.749) \\

\hline
\noalign{\vskip 1pt}
\rowcolor{gray!15}
Expert
& 0.711 (0.669, 0.750) \\

Expert + ResNet-18
& 0.735 (0.694, 0.774)$^{*}$ \\

Expert + ResNet-18+2zones
& 0.747 (0.705, 0.785)$^{*}$ \\

Expert + ResNet-18+5zones
& \textbf{0.756 (0.716, 0.793)}$^{*}$ \\

Expert + ResNet-50
& 0.745 (0.703, 0.782)$^{*}$ \\

Expert + ResNet-50+2zones
& \textbf{0.756 (0.714, 0.795)}$^{*}$ \\

Expert + ResNet-50+5zones
& 0.749 (0.708, 0.786)$^{*}$ \\

\hline

ResNet-50+5zones, csPCa
& \textbf{0.809 (0.782, 0.834)} \\

\hline
\end{tabular}%
}
\end{table}

\subsection{External evaluation}
Table \ref{tab:external} summarizes performance on the UCSF and PMUB datasets. 
Fine-tuning models pretrained on the internal cohort consistently outperformed training from scratch, indicating that representations learned from the large internal dataset provided a useful initialization for external adaptation.
Anatomical guidance continued to provide benefits across both datasets.
On UCSF, 5-zone guidance significantly improved ResNet-50 PI-RADS prediction (QWK: -0.005 to 0.231; $p<0.05$), while ResNet-18+2-zones achieved the highest csPCa AUC (0.668). 
On PMUB, ResNet-18+2-zones achieved the highest PI-RADS QWK (0.368), whereas ResNet-18+5-zones achieved the highest csPCa AUC (0.683).
These results suggest that the benefits of zonal information generalize beyond a single training cohort.

\subsection{Clinical utility of model-predicted PI-RADS scores}
We compared csPCa detection based on expert-assigned PI-RADS scores, model-predicted PI-RADS scores, and their combination (Table~\ref{tab:pirad-cspca}). 
While ProFound-Conv and ViT underperformed expert PI-RADS assessment, anatomy-guided ResNet models produced PI-RADS predictions that were at least comparable to expert scoring for csPCa discrimination. The best model-predicted PI-RADS score (ResNet-50+2-zones) achieved an AUC of 0.729, exceeding that of expert-assigned PI-RADS scores (0.711). Combining expert and model-predicted PI-RADS scores yielded the strongest performance (AUC = 0.756, $p<0.05$), indicating that the two sources of information provide complementary value.

\section{Conclusion}
\label{sec:Conclusion}


We evaluated the impact of anatomical guidance on DL models for PI-RADS prediction and csPCa detection.
Across architectures trained using an internal dataset and fine-tuned using external datasets, 
incorporating prostate zonal masks generally improved performance.
Anatomical guidance improved the performance of model-predicted PI-RADS scores and csPCa detection, and enhanced PI-RADS-based detection of csPCa.
Combining expert-assigned and model-predicted PI-RADS scores improved PI-RADS-based csPCa detection, suggesting that AI-derived PI-RADS scores may provide complementary information for csPCa detection. 
These findings highlight the value of incorporating anatomical context into automated prostate MRI assessment.

This study has several limitations. 
First, segmentation errors in anatomical guidance may propagate to downstream tasks. 
Second, although external validation was performed on two independent datasets, both are limited to only PI-RADS 3–5 cases, limiting evaluation across the full PI-RADS spectrum. 
Future work should investigate whether the benefits of anatomical guidance generalize across more datasets and downstream clinical tasks.

\section{Compliance with ethical standards}
\label{sec:ethics}

This retrospective study was approved by the Institutional Review Board (IRB), with a waiver of informed consent, and was conducted in compliance with the Health Insurance Portability and Accountability Act (HIPAA).

\section{Acknowledgments}
\label{sec:acknowledgments}

This work was partly supported by the National Institutes of Health grants K25CA283145, R01AG053949, R37CA282407, R01AG064027, and R01AG070988.
Mert R. Sabuncu serves as a consultant or advisor to Cleerly Inc. and Medara Health, and holds equity interests. 
Daniel Margolis has served as a clinical advisor to Stratagen Bio and Ravada Health and holds unexercised stock options.
Claude Code was used for partial code implementation. 


\bibliographystyle{IEEEbib}
\bibliography{strings}

\end{document}